\documentclass[aapm,mph,preprint,graphicx]{revtex4-1}

\usepackage{graphicx}
\usepackage{multirow}
\usepackage{amsfonts,amsmath, amsbsy, amssymb}
\usepackage{epsfig}
\usepackage{hyperref}
\usepackage[displaymath,mathlines]{lineno}
\usepackage{algorithmicx,algpseudocode}

\newcommand{\Lc}{{\mathcal L}}

\newcommand{\Ed}{{\mathbb{E}}}

\begin{document}

\title{Cycle Consistent Adversarial Denoising Network for Multiphase Coronary CT Angiography}

\author{Eunhee Kang$^{1}$, Hyun Jung Koo$^{2}$, Dong Hyun Yang$^{2}$, Joon Bum Seo$^{2}$ and Jong Chul Ye$^{1*}$}

\affiliation{$^{1}$Bio Imaging and Signal Processing Lab., Dept. of Bio and Brain Engineering, KAIST, Daejeon, Republic of Korea\\
			 $^{2}$Dept. of Radiology, Asan Medical Center, University of Ulsan College of Medicine, Seoul, Republic of Korea\\
			 $^{*}$Corresponding author: jong.ye@kaist.ac.kr}

\begin{abstract}

\textbf{Purpose:}
In multiphase coronary CT angiography (CTA), a series of CT images are taken at different levels of radiation dose during the examination. Although this reduces the total radiation dose, the image quality during the low-dose phases is significantly degraded. Recently, deep neural network approaches based on supervised learning technique have demonstrated impressive performance improvement over  conventional model-based iterative methods for low-dose CT.
However,   matched low- and routine- dose CT image pairs are difficult to obtain in multiphase CT.
To address this  problem, we aim at developing a new deep learning framework. 

\noindent\textbf{Method:}
We propose an unsupervised learning technique that can remove the noise of the CT images in the low-dose phases by learning from the CT images in the routine dose phases.
Although a supervised learning approach is not applicable due to the differences in the underlying heart structure in two phases, the images  are closely related  in two phases, so we propose a cycle-consistent adversarial denoising network to learn the  mapping between the low and high dose cardiac phases.

\noindent\textbf{Results:}
Experimental results showed that the proposed method effectively reduces the noise in the low-dose CT image while  preserving detailed texture and edge information.
Moreover, thanks to the cyclic consistency and identity loss, the proposed network does not create any artificial features that are not present in the input images.
Visual grading and quality evaluation also confirm that the proposed method provides significant improvement in diagnostic quality.

\noindent\textbf{Conclusions:}
The proposed network can learn the image distributions from the routine-dose cardiac phases, which is a big advantages over
the existing supervised learning networks that need exactly matched low- and routine- dose CT images.
Considering the effectiveness and practicability of the proposed method, we believe that the proposed can be applied for many other CT acquisition protocols.

\end{abstract}

\maketitle


\section{Introduction}

X-ray computed tomography (CT) is one of the most widely used imaging  modalities for diagnostic purpose.
For example, in cardiac coronary CT angiography (CTA), a series of CT images are acquired while examination is conducted with contrast injection \cite{hsieh2009computed}, 
which helps clinicians to identify  heart disease. 
However, it is often difficult to predict in which heart phase the disease area can be seen better,
so multiphase acquisition is often necessary. 
In the case of valve disease, cardiac motion information from multiphase acquisition is essential, and
in evaluating cardiac function, myocardial motion should be evaluated, in which case multiphase acquisition is needed.

\begin{figure}[b]
\centering
\includegraphics[width=10cm]{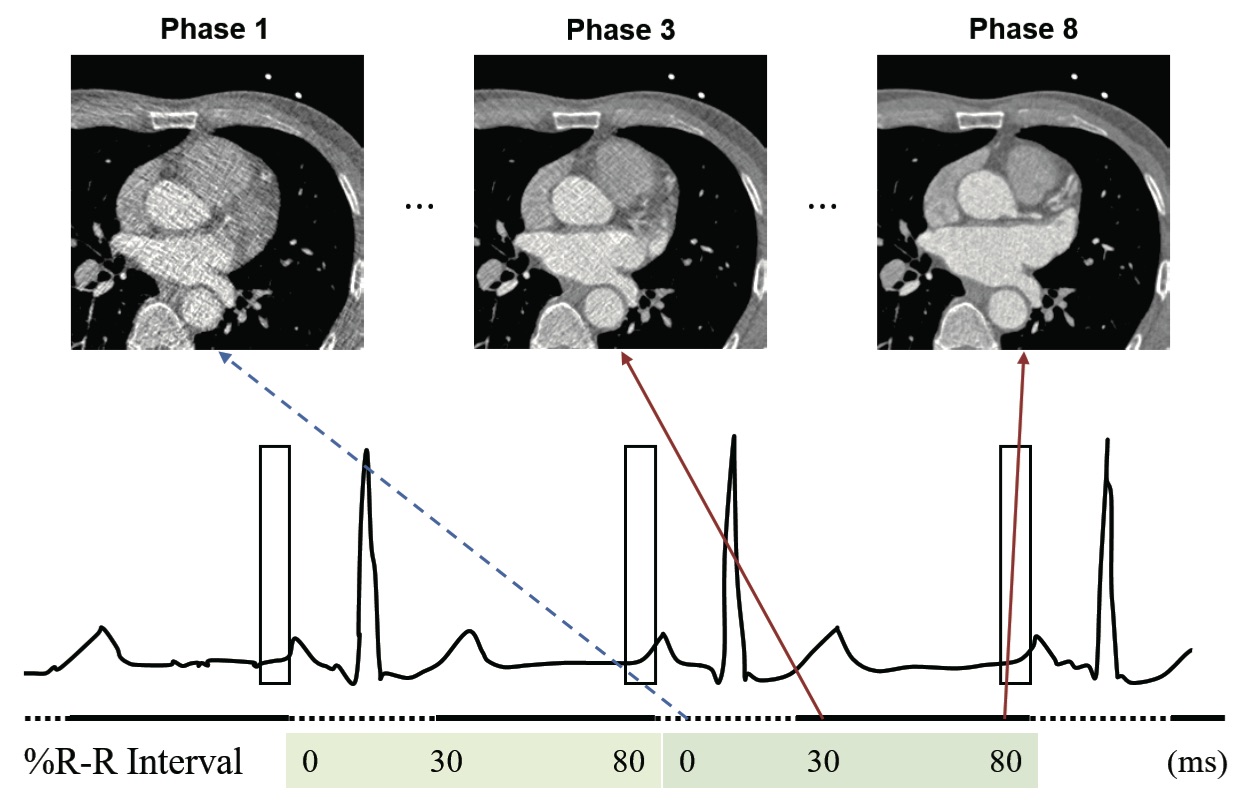}
\caption{Example of multiphase coronary CTA acquisition protocol.
Low-dose acquisition is performed in phase 1 and 2,  whereas routine-dose acquisition is done in phase from 3 to 10.}
\label{fig:cardiac_CT_example}
\end{figure}

However, taking all phase images at full dose is  not allowed due to  the excessive radiation dose. On the other hand, it is risky to get the entire phase at the low dose because none of the cardiac phases may be diagnostically useful. Therefore, in clinical environments, a multiphase tube current modulated CTA  as shown in Fig. \ref{fig:cardiac_CT_example} is often used to obtain at least one high-dose phase image, which information may also be exploited by the radiologist to interpret the low-dose phase images.

Although this tube current modulation can reduce the total radiation dose, it also  introduces noise in the projection data of the low-dose phases.
This results in CT images with different noise levels and contrast at different cardiac phases (see Fig. \ref{fig:cardiac_CT_example}).
Although model-based iterative reconstruction (MBIR) methods \cite{beister2012iterative,ramani2012splitting,sidky2008image,leipsic2010adaptive,funama2011combination,chen2008prior,renker2011evaluation} have been developed to address this,
the MBIR approaches suffer from relatively long reconstruction time due to the iterative applications of forward and back projections.

Recently, deep learning approaches have demonstrated impressive performance improvement over conventional iterative methods for low-dose CT \cite{kang2017deep,kang2018framelet,chen2017low,yang2017ct} and sparse-view CT \cite{ye2017deep,han2018sparse,jin2017sparse,adler2017learned}.
The main advantage of deep learning approach over the conventional MBIR approaches is that the network learns the image statistics in a fully data-driven way rather than using hand-tuned regularizations.
While these approaches usually take time for training, real-time reconstruction is possible once the network is trained, making the algorithm very practical in the clinical setting.


While these networks  have been designed based on supervised learning technique,
in real clinical situation matched low- and routine- dose CT image pairs are difficult to obtain.
The matched full/low dose data are only available when 1) additional full-dose acquisition is available, or 2) simulated low-dose data can be generated from the full-dose acquisition. However, multiple acquisition at different doses is not usually allowed for human study due to the additional radiation dose to patients. Even when such experiments are allowed, the multiple acquisitions  are usually associated with motion artifacts due to patients and gantry motions. Therefore, most of the current work uses the simulated low-dose data provided by the vendors (for example, AAPM  (American Association of Physicists in Medicine) low-dose grand challenge data set \cite{mccollough2017low}).  However, in order to have realistic low-dose images, noise should be added in the sinogram domain, so independent algorithm development without vendor assistance is very difficult. Moreover, there are concerns that the noise patterns in the simulated low-dose image is somewhat different from real low-dose acquisition, so the supervised learning with simulated data can be biased.

To address this unmatched pair problems, Wolterink et al \cite{wolterink2017generative} proposed a low-dose CT denoising network with generative adversarial network (GAN) loss so that the distribution of the denoising network outputs can match the routine dose images.
However, one of the important limitations of GAN for CT denoising \cite{yang2018lowgan,Yi2018gan,wolterink2017generative} is that there is a  risk that the network may generate features that are not present in the images due to the potential mode-collapsing behavior of GAN. 
The GAN mode collapse occurs when the generator network generates limited outputs or even the same output, regardless of the input \cite{arjovsky2017wasserstein,zhu2017cyclegan}.
This happens when GAN is trained to match the data distributions but it does not guarantee that an input and output data are paired in a meaningful way, i.e.
there can exist various inputs (resp. outputs) that matches to one output (resp. input) despite them generating the same distribution.


In coronary CTA,  
even though the images at the low-dose and high-dose phases do not match each other exactly due to the cardiac motion,
they are from the same cardiac volume so that they have important correspondence.
Therefore, one can conjecture that the correctly denoised low-dose phase should follow the routine dose phase image distribution more closely 
and learning between two phase cardiac images is more effective than learning from totally different images.
One of the most important contributions of this work is to show that we can indeed 
improve the CT images at the low-dose phase by learning the distribution of the images at the high-dose phases using the {\em  cyclic consistency} by Zhu et al. (cycle GAN) \cite{zhu2017cyclegan} or by Kim el al. (DiscoGAN) \cite{kim2017discogan}.
Specifically, we train two networks between  two different domains (low dose and routine dose).
Then, the training goal is that the two networks should be inverse of each other.
Thanks to the existence of inverse path that favors the one to one correspondence between the input and output, the training of the GAN  is less affected by the mode collapse.
Furthermore, unlike the classic GAN which generates samples from random noise inputs, our network creates samples from the noisy input that are closely related. 
This also reduces the likelihood of mode collapse.
Another important aspect of the algorithm is  the identity loss \cite{zhu2017cyclegan}.  
The main idea of the identity loss is that a generator  $G: A\mapsto B$ should work as an identity
for the target domain image $y\in B$ such that $G(y)\simeq y$.  This constraint 
works as a fixed-point constraint of the output domain so that as soon as the output signal is generated to match the target distribution, the network no longer changes the signal.
Experimental results show that the proposed method is robust to the cardiac motion and contrast changes and does not create artificial features.

\section{Theory}
\label{sec:method}

%

\begin{figure}[t!]
\centering
\includegraphics[width=13cm]{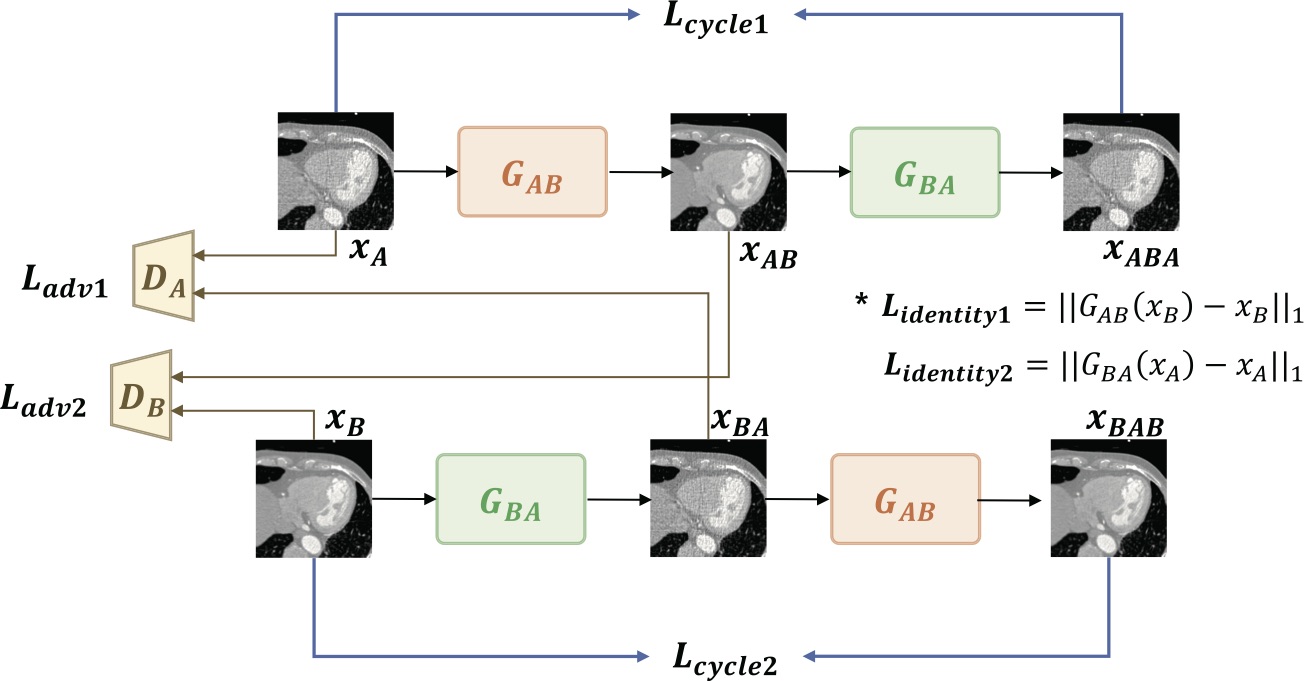}
\caption{Overview of the proposed framework for low-dose CT image denoising.
There are two generator networks $G_{AB}$ and $G_{BA}$ and two discriminator networks $D_A$ and $D_B$.
$A$ denotes the low-dose CT image domain and $B$ denotes the routine-dose CT image domain.
The network employes three losses such as adversarial loss (adv), cyclic loss, and additionally identity loss.}
\label{fig:framework}
\end{figure}

The overall framework of the proposed network architecture is illustrated in Fig. \ref{fig:framework}.
We denote the  low-dose CT domain  by $(A)$ and routine-dose CT domain by $(B)$, and
the probability  distribution for each domain is referred to as $P_A$ and $P_B$, respectively.
The generator $G_{AB}$  denotes the mapping from $(A)$ to $(B)$, and $G_{BA}$ are similarly defined as the mapping from $(B)$ to $(A)$. 
As for the generator, we employ the optimized network for a noise reduction in low-dose CT images in our prior work\cite{kang2018framelet}.
In addition, there are two adversarial discriminators $D_A$ and $D_B$ which distinguish between measured input images and synthesized images from the generators.
Then, we train the generators and discriminators simultaneously.
Specifically, we aim to solve the following optimization problem:
\begin{eqnarray}
\min_{G_{AB},G_{BA}} \max_{D_A,D_B} \Lc(G_{AB},G_{BA},D_A, D_B).
\end{eqnarray}
where the overall loss is defined by:
\begin{align}
\Lc(G_{AB},G_{BA},D_A,D_B) =& \Lc_{GAN}(G_{AB},D_B,A,B) + \Lc_{GAN}(G_{BA},D_A,B,A) \nonumber \\
						   +& \lambda \Lc_{cyclic}(G_{AB},G_{BA}) + \gamma \Lc_{identity}(G_{AB},G_{BA}) ,
\label{eq:loss}
\end{align}
where $\lambda$ and $\gamma$ control the importance of the losses,
and $\Lc_{GAN}$, $\Lc_{cyclic}$ and $\Lc_{identity}$ denote the adversarial loss, cyclic loss, and identity loss. 
More detailed description of each loss follows.

\subsection{Loss formulation}

\subsubsection{Adversarial loss}

We employ adversarial losses using GAN as proposed in Zhu et al\cite{zhu2017cyclegan}.
According to the original GAN \cite{goodfellow2014gan}, the generator $G_{AB}$ and discriminator $D_B$ can be trained by solving the following min-max problem:
\begin{align}
\min_{G_{AB}} \max_{D_B} \Lc_{GAN}(G_{AB},D_B,A,B) = \Ed_{x_B\sim P_B} [\log D_B(x_B)] + \Ed_{x_A\sim P_A} [\log(1-D_B(G_{AB}(x_A)))],
\label{eq:original_gan_loss}
\end{align}
where $G_{AB}$ is trained to reduce a noise in the low-dose CT image $x_A$ to make it similar to the routine-dose CT image $x_B$,
while $D_B$ is trained to discriminate between the denoised CT image $G_{AB}(x_A)$ and the routine-dose CT image $x_B$.
However, we found that the original adversarial loss \eqref{eq:original_gan_loss} is unstable during training process; 
thus, we changed the log-likelihood function to a least square loss as in the least squares GAN (LSGAN) \cite{mao2017least}. 
Then, the min-max problem can be changed to the two minimization problems as follows:
\begin{align}
\min_{G_{AB}} & \Ed_{x_A \sim P_A} [(D_B(G_{AB}(x_A)) - 1)^2], \label{eq:loss_generator}\\
\min_{D_B} & \frac{1}{2}\Ed_{x_B \sim P_B} [(D_B(x_B) - 1)^2] + \frac{1}{2}\Ed_{x_A \sim P_A} [D_B(G_{AB}(x_A))^2]. \label{eq:loss_discriminator}
\end{align}
The adversarial loss causes the generator to generate the denoised images  that may deceive the discriminator to classify them as the real images at routine doses.
At the same time, the adversarial loss will guide the discriminator to well distinguish the denoised image and the routine dose image.
Similar adversarial loss is added to the generator $G_{BA}$, which generates noisy images.

\subsubsection{Cyclic loss}

With the adversarial losses, we could train the generator $G_{AB}$ and $G_{BA}$ to produce the realistic denoised images and noisy CT images, respectively;
but this does not guarantee that they have an inverse relation described in Fig. \ref{fig:framework}.
To enable one to one correspondence between the noisy and denoised image, 
the cycle which consists of two generators should be imposed to bring the input $x_A$ to the original image.
More specifically, the cyclic loss is defined by
\begin{align}
\Lc_{cyclic}(G_{AB},G_{BA}) = \Ed_{x_A \sim P_A} [\|G_{BA}(G_{AB}(x_A))-x_A\|_1] + \Ed_{x_B \sim P_B} [\|G_{AB}(G_{BA}(x_B))-x_B\|_1],
\end{align}
where $\|\cdot\|_1$ denotes the $l_1$-norm.
Then,  the cyclic loss enforces the constraint that
$G_{AB}$ and $G_{BA}$ should be inverse of each other, i.e.  it encourages 
$G_{BA}(G_{AB}(x_A)) \approx x_A$ and $G_{AB}(G_{BA}(x_B)) \approx x_B$.

\subsubsection{Identity loss}

In multiphase CTA, there are often cases where the heart phase and dose modulation are not perfectly aligned as originally planned. 
 For example, in the multiphase CTA acquisition in Fig.~\ref{fig:cardiac_CT_example},
it is assumed that the systolic phase images should be obtained using low dose modulation, but due to the mismatch with the cardiac cycle from arrhythmia, the systolic phase image noise level may vary and even be in full dose.
In this case,  the input to the generator $G_{AB}$
can be at full dose, so  it is important to train the generator so that it does not
alter such clean images.
Similarly, the generator $G_{BA}$ should not change the input images acquired at the low-dose level.
To enforce the two generator $G_{AB}$ and $G_{BA}$ to satisfy these conditions,
the following identity loss should be minimized: 
\begin{align}\label{eq:idcond}
\Lc_{identity}(G_{AB},G_{BA}) = \Ed_{x_B \sim P_B} [\|G_{AB}(x_B)-x_B\|_1] + \Ed_{x_A \sim P_A} [\|G_{BA}(x_A)-x_A\|_1].
\end{align}
In other word, the generators should work as identity mappings for the input images at the target domain:
\begin{eqnarray}\label{eq:idour}
G_{AB}(x_B) \simeq x_B,\quad G_{BA}(x_A) \simeq x_A
\end{eqnarray}
Note that this identity loss is  similar to 
the  identity loss for the photo generation from paintings   in order to maintain the color composition between input and output domains\cite{zhu2017cyclegan}.
The constraints in \eqref{eq:idour} ensure 
 that the correctly generated output images no longer vary when used as inputs to the same network, i.e.
 the target domain should be the fixed points of the generator.
 As will be shown later in experiments, this constraint is important to avoid creating artificial features.

\begin{figure}[!hbt]
\centering
\includegraphics[width=11cm]{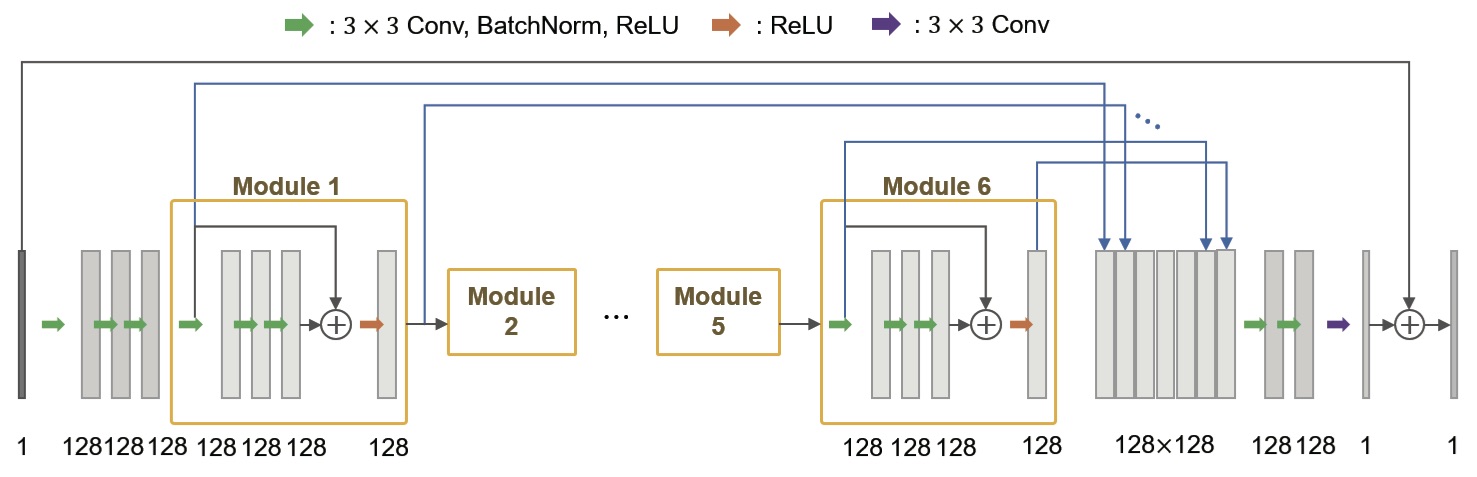}
\caption{A generator architecture optimized for the low-dose CT image denoising \cite{kang2018framelet}.}
\label{fig:generator}
\end{figure}

\begin{figure}[!hbt]
\centering
\includegraphics[width=10cm]{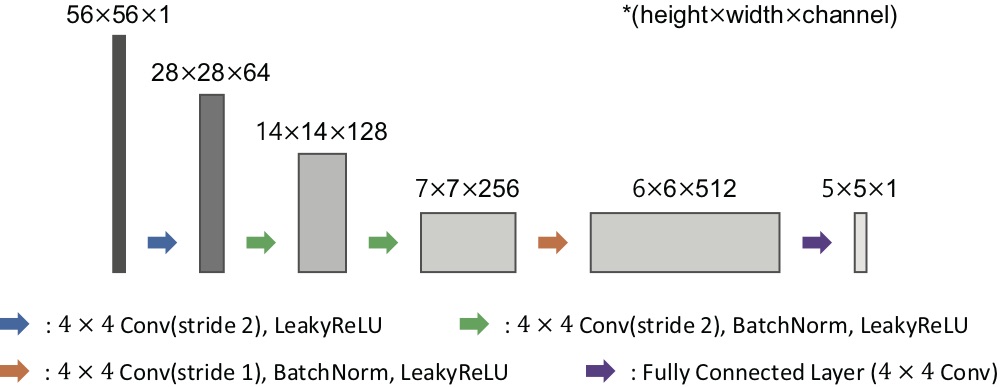}
\caption{A network architecture of discriminator  \cite{isola2017image}.}
\label{fig:discriminator}
\end{figure}

\subsection{Network architecture}

The network architecture of two generators $G_{AB}$ and $G_{BA}$ is illustrated in Fig. \ref{fig:generator}.
This architecture is optimized for low-dose CT image denoising in Kang et al\cite{kang2018framelet}.
To reduce network complexity, images are used directly as inputs to the network instead of the wavelet transform coefficients as in our prior work\cite{kang2018framelet}.
The first convolution layer uses 128 set of $3 \times 3$ convolution kernels to produce 128 channel feature maps.
We have 6 set of module composed of 3 sets of convolution, batch normalization, and ReLU layers, and one bypass connection with a ReLU layer.
Convolution layers in the modules use 128 set of $3 \times 3 \times 128$ convolution kernels.
In addition, the proposed network has a concatenation layer that concatenates the inputs of each module and the output of the last module, 
which is followed by the convolution layer with 128 set of $3 \times 3 \times 896$ convolution kernels.
This concatenation layer has a signal boosting effect using multiple signal representation \cite{kang2018framelet} 
and provides various paths for gradient backpropagation.
The last convolution layer uses 15 sets of $3 \times 3 \times 128$ convolution kernels.
Finally, we add an end-to-end bypass connection to estimate the noise-free image while exploiting the advantages of bypass connection in He et al\cite{he2016deep}.

The network architecture of discriminators $D_A$ and $D_B$ is illustrated in Fig. \ref{fig:discriminator}.
This is from PatchGAN \cite{isola2017image}, which has $ 70 \times 70$ receptive field and classifies image patches whether they are real or synthesized.
Specifically, it consists of 5 convolution layers including the last fully-connected layer.
The first convolution layer uses 64 sets of $ 4 \times4 $ convolution kernels, and the number of convolution kernels in the following layers is twice that of the previous layer except the last fully connected layer.
After the last fully connected layer,  $5\times 5$ feature maps are obtained, and we calculate the
$l_2$-loss.
Arbitrary sized images can be applied to this discriminator network by summing up the $l_2$-loss from each $56\times 56$ patch, after which
the final decision is made.

\section{Methods}

\subsection{Data: Cardiac CT scans}

The study cohort comprised 50 CT scans of mitral valve prolapse patients and 50 CT scans of coronary artery disease patients,  and the CT scan protocols are described in previous reports \cite{koo2014Demonstration, yang2015stress}. 
The mean age of the population was $58 \pm 13.2$ years, and the mean body weight was $66.2 \pm 12.6$ kg. 
Using a second generation dual source CT scanner (Somatom Definition Flash, Siemens, Erlangen, Germany), electrocardiography (ECG)-gated cardiac CT scanning was performed. 
Retrospective ECG-gated spiral scan with ECG-based tube current modulation was applied to multiphase of 0-90\% of the R-R interval which comprises with a full dose pulsing window of 30-80\% of the R-R interval. 
The tube current was reduced to 20\% of the maximum outside the ECG pulsing window \cite{weustink2008optimal} (Fig. \ref{fig:cardiac_CT_example}).
A bolus of 70-90 mL of contrast material (Iomeprol, Iomeron 400; Bracco, Milan, Italy) was administered by a power injector (Stellant D; Medrad, Indianola, PA, USA) at a rate of 4.0 mL/s and followed by 40 mL saline.
The bolus tracking method (region of interest, the ascending aorta; attenuation threshold level, 100 HU; scan delay, 8 s) was applied to determine scan time. 
In all CT scans, tube voltage and the tube current\textendash exposure time product were adjusted according to the patients body size, and the scan parameters were as follows: 
tube voltage, 80-120 kV; tube current\textendash exposure time product, 185-380 mAs; collimation, $128 \times 0.6$ mm; and gantry rotation time, 280 s. Mean effective radiation dose of CCTA was $11.4 \pm 6.2$ mSv. 
Standard cardiac filter (B26f) was used for imaging reconstruction.

\subsection{Training details}

Training was performed by minimizing the loss function (\ref{eq:loss}) with $\lambda=10$ and $\gamma=5$.
We used the ADAM optimization method to train all networks with $\beta_1=0.5$ and $\beta_2=0.999$.
The number of epochs was 160, which was divided  into two phases to control the learning rate during the training.
In the first 100 epochs, we set the learning rate to 0.0002, and linearly decreased it to zero over the next epochs.
We performed early stopping at 160 epochs, since the early stopping was shown to work as a regularization \cite{caruana2001overfitting}.
The size of patch was $56\times56$ and the size of mini-batch was 10.
Kernels were initialized randomly from a Gaussian distribution.
We have updated the generator and the discriminator at each iteration.
We normalized the intensity of the input low-dose CT images and the target routine-dose CT image using the maximum intensity value of the input images, and subtract 0.5 and multiply two to make the input image intensity range as $[-1,1]$.
For training, we used 50 cases from the dataset of mitral valve prolapse patients.
The proposed method was implemented in Python with the PyTorch \cite{paszke2017automatic} 
and NVIDIA GeForce GTX 1080 Ti GPU was used to train and test the network.

\subsection{Evaluation}

\subsubsection{Visual grading analysis}

Image quality was assessed using relative visual grading analysis (VGA). 
This VGA method is planned to be related to the clinical task to evaluate any structural abnormality that may present at specific anatomical structures in a CT images. 
Two expert cardiac radiologists established a set of anatomical structures to evaluate image quality. 
Table \ref{table1} demonstrates the 13 anatomical structures used in this study. 
The VGA scoring scale are shown in Table \ref{table2}. 
All CT images including denoising CT images were uploaded on picture archiving and communication system (PACS) for visual grading. 
Of all, randomly selected 25 CT scans from mitral valve prolapse patients and 25 CT scans from coronary artery disease patients were included for VGA. 
{Total 1300 CT images (50 selected CT scans $\times$ 13 structures $\times$ original and denoising CT) were scored. }
Two radiologists performed VGA in consensus, and all CT scans are scored independently, without side-by-side comparison.

\begin{table}[t]
\caption{Structures selected as diagnostic requirements to assess the diagnostic quality of cardiac images. The structures were evaluated to be sharp with clear visualization. (LCA, left coronary artery; LV, left ventricle; RCA, right coronary artery; RV, right ventricle)}
\label{table1}
\centering
\resizebox{0.33\paperwidth}{!}{
\begin{tabular}{l|l}
\hline
Organ 							& Structure \\ \hline\hline
Left/right coronary artery		& LCA ostium \\ 
								& LCA distal 1.5 cm \\ 
								& LCA distal \\ 
								& RCA ostium \\ 
								& RCA 1.5 cm \\ 
								& RCA distal \\ 
Cardiac wall 					& LV septum \\
								& RV free wall margin \\
Cardiac cavity 					& LV trabeculation \\
								& Left arterial appendage \\
Aorta 							& Aortic root \\
Valve 							& Aortic valve \\
								& Mitral valve \\ \hline
\end{tabular}}
\end{table}

\begin{table}[t]
\caption{Visual grading analysis scores used to evaluate the structure visibility}
\label{table2}
\centering
\resizebox{0.33\paperwidth}{!}{
\begin{tabular}{c|l}
\hline
\multirow{2}*{Score} & Visibility of the structures in relation to \\
 		& the reference images \\ \hline
1		& Poor image quality \\
2 		& Lower image quality \\
3 		& Mild noise, but acceptable \\
4 		& Average \\
5 		& Good \\
6 		& Excellent \\ \hline
\end{tabular}}
\end{table}

\subsubsection{Quantitative analysis}

The image noise and signal-to-noise (SNR) of all images were obtained at four anatomical structures: ascending aorta, left ventricular cavity, left ventricular septum, and proximal right coronary artery. 
The size of region of interest to evaluate SNR were varied to fit each anatomic structure; however, it was confined into each structure without overlapping other structures.

\subsection{Statistical analysis}

VGA scores obtained from original CT images and denoising images were compared using chi-square test. 
Image noise and SNR were compared using paired $t$-test. 
$P$-values of $<0.05$ indicated statistical significance. Statistical analyses were performed using commercial software (SPSS, Chicago, IL, United States).

\begin{figure*}[t!]
\centering
\includegraphics[width=17cm]{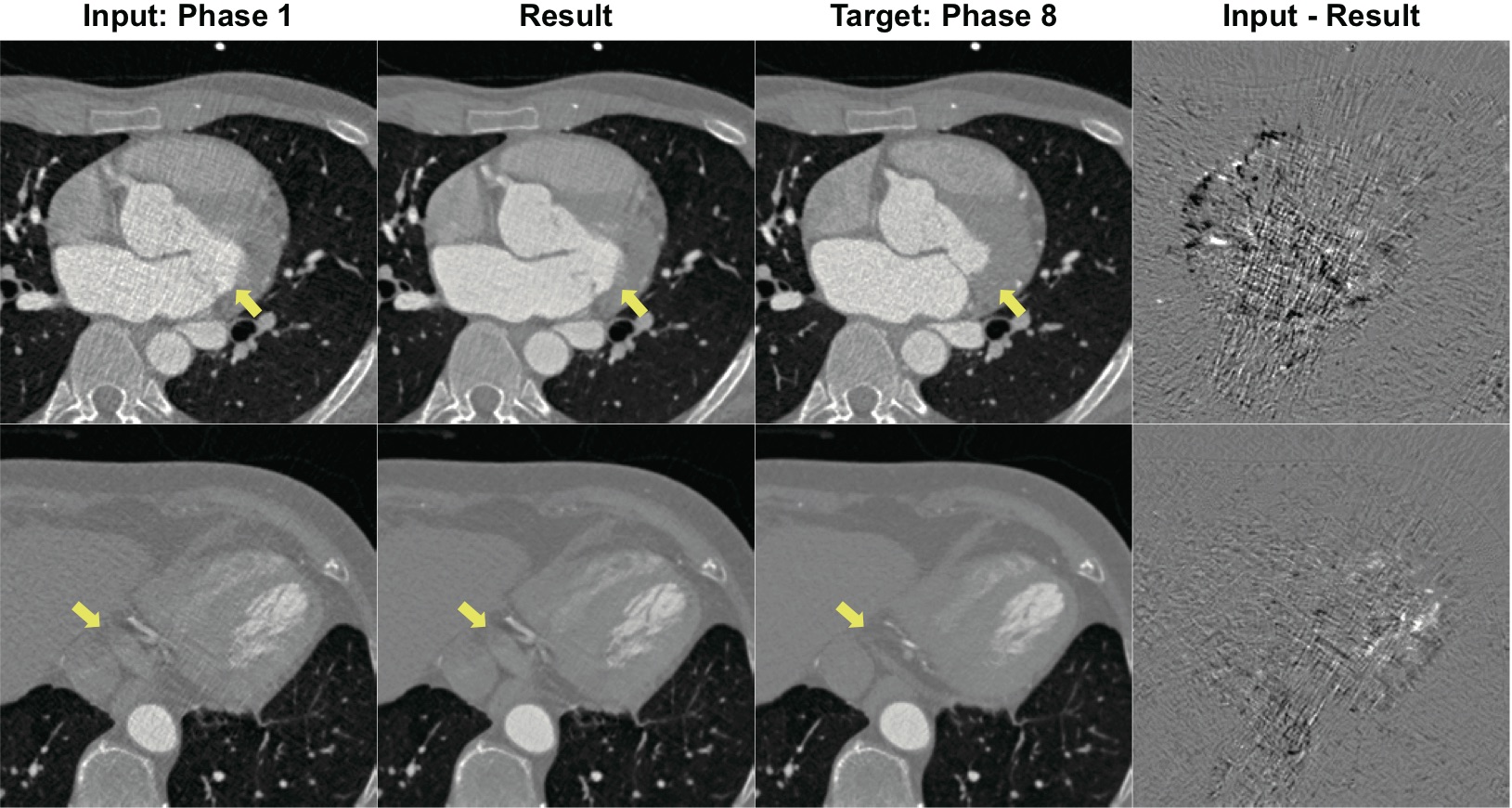}
\caption{Restoration results from the dataset of mitral valve prolapse patients.
Intensity range of the CT image is (-1024, 976)[HU] and the difference image between the input and result is (-150, 150)[HU].
Yellow arrow indicates the distinctly different region between input image from phase 1 and target image from phase 8.}
\label{fig:result_1}
\end{figure*}

\begin{figure*}[t!]
\centering
\includegraphics[width=17cm]{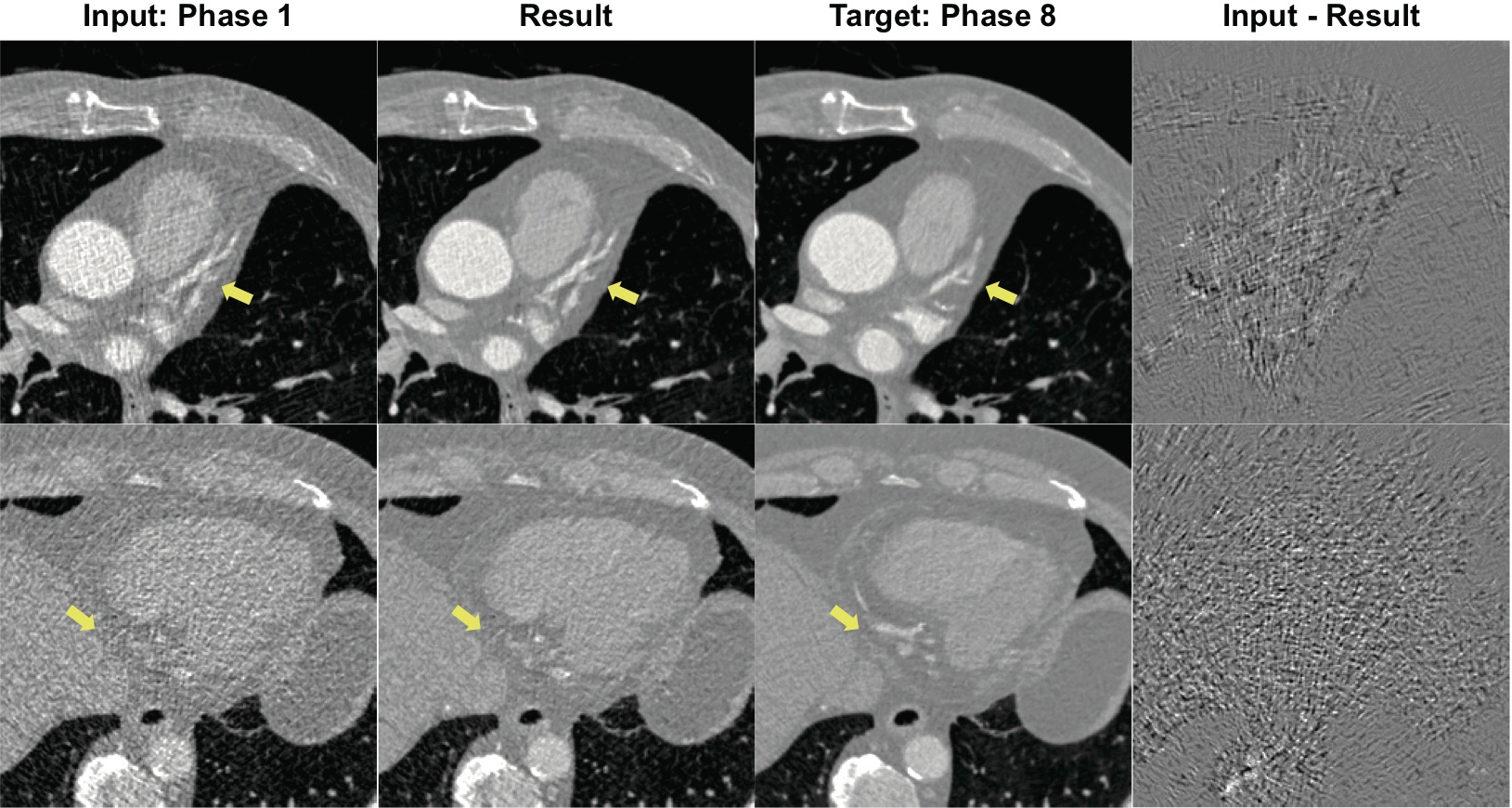}
\caption{Restoration results from the dataset of coronary artery disease patients.
Intensity range of the CT image is (-924, 576)[HU] and the difference image between the input and result is (-200, 200)[HU].
Yellow arrow indicates the distinctly different region between input image from phase 1 and target image from phase 8.}
\label{fig:result_2}
\end{figure*}

\begin{figure*}[t!]
\centering
\includegraphics[width=17cm]{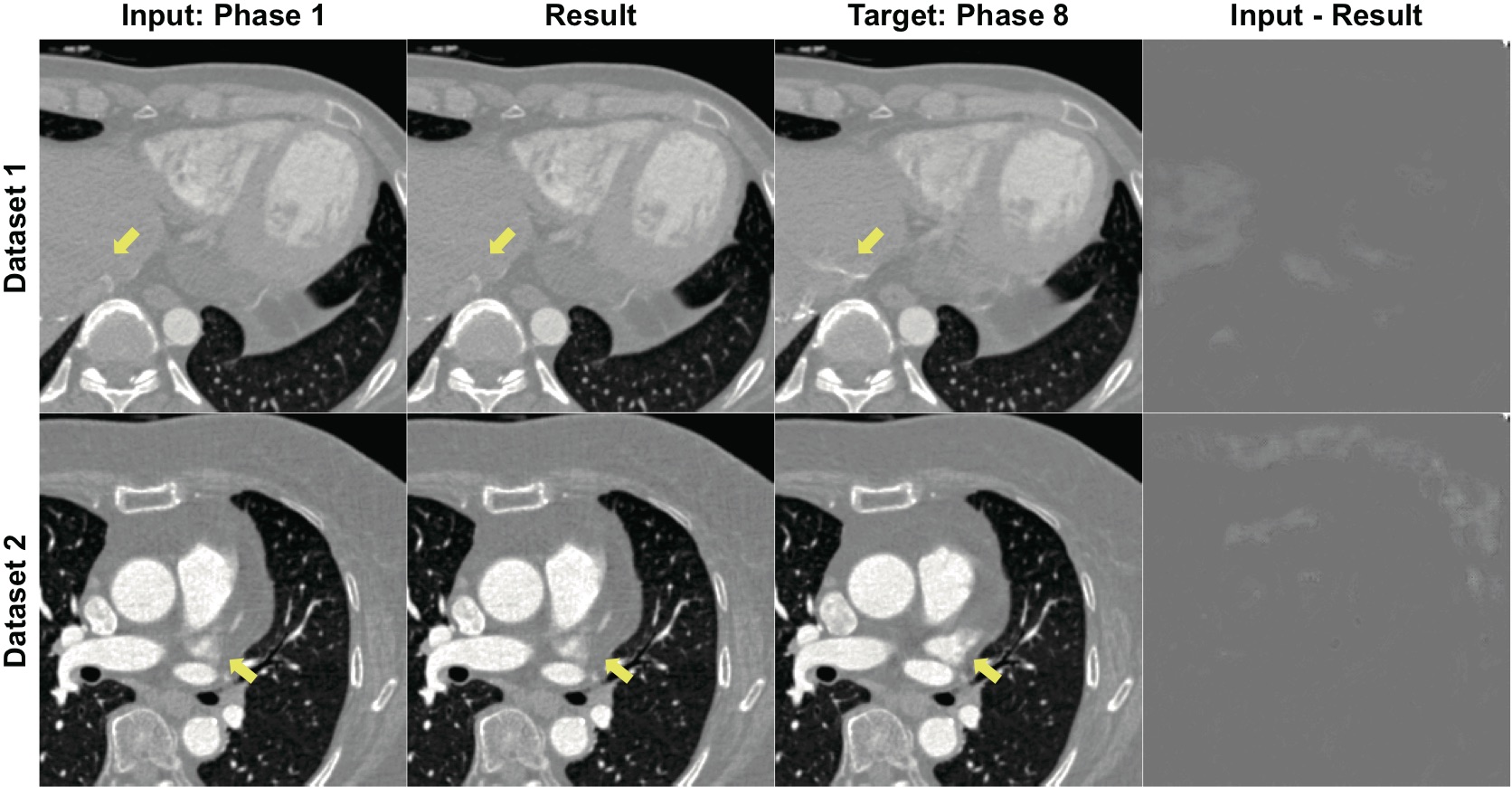}
\caption{Restoration results from the dataset whose input CT images have similar noise level with the target CT images.
Intensity range of the CT image is (-924, 576)[HU] and the difference image between the input and result is (-15, 15)[HU].
Yellow arrow indicates the distinct different region between input image from phase 1 and target image from phase 8.}
\label{fig:result_identity}
\end{figure*}

\begin{figure*}[t!]
\centering
\includegraphics[width=12cm]{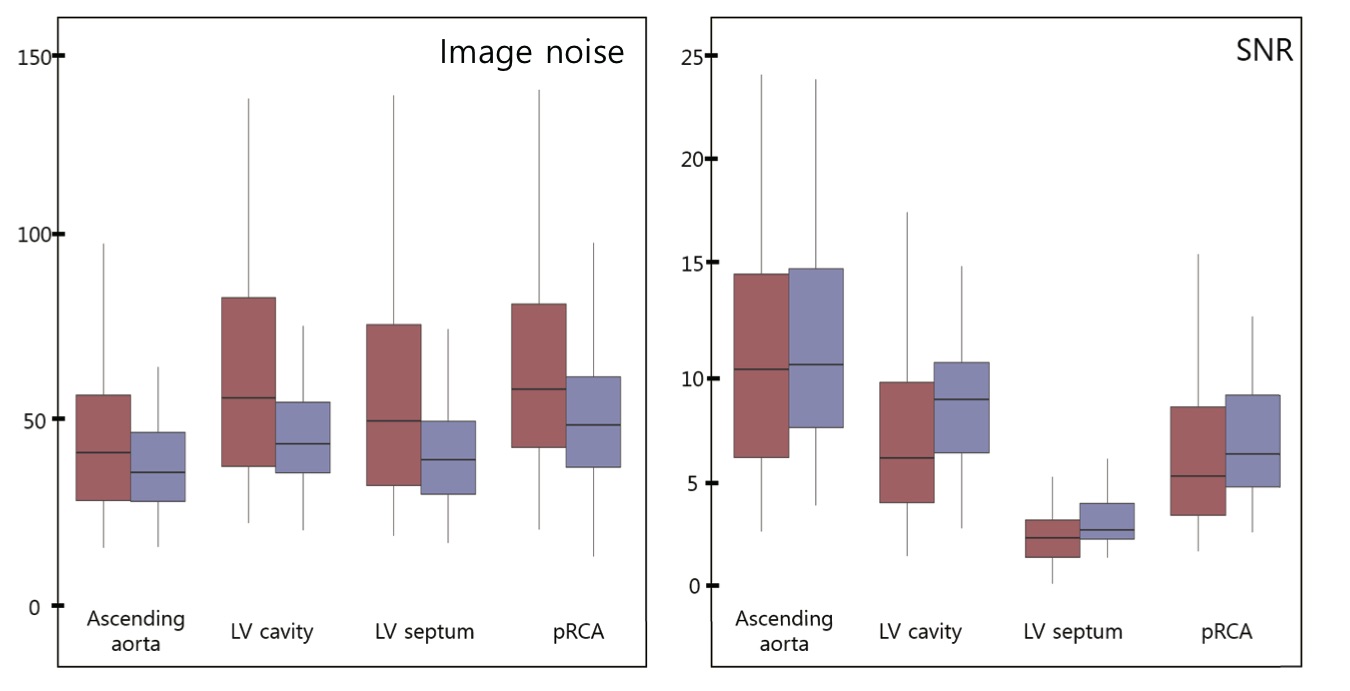}
\caption{Standard deviation and signal-to-noise ratio between original CT (red) and denoising CT (purple) images measured from selected structures. (LV, left ventricle; pRCA, proximal right coronary artery)}
\label{fig:result_noise_SNR}
\end{figure*}

\begin{table*}[t!]
\caption{Comparison of visual scores between original image and denoising CT image using Chi square method.
(LCA, left coronary artery; LV, left ventricle; RCA, right coronary artery; RV, right ventricle)}
\label{table3}
\centering
\resizebox{0.55\paperwidth}{!}{
\begin{tabular}{llllc}
\hline
Organ 							& Structure 		& Original image & Denoising 		& P-value \\ \hline
Left/right coronary artery		& LCA ostium		& $3.5 \pm 1.4$  & $4.2 \pm 1.1$ 	& $<0.001$\\ 
								& LCA distal 1.5 cm & $3.4 \pm 1.4$  & $4.1 \pm 1.1$ 	& $<0.001$\\  
								& LCA distal 		& $2.3 \pm 1.3$  & $3.0 \pm 1.2$ 	& $<0.001$\\  
								& RCA ostium		& $3.3 \pm 1.3$  & $4.0 \pm 1.0$ 	& $<0.001$\\  
								& RCA 1.5 cm 		& $3.1 \pm 1.6$  & $3.9 \pm 1.1$ 	& $<0.001$\\  
								& RCA distal 		& $2.3 \pm 1.4$  & $2.8 \pm 1.2$ 	& $<0.001$\\  
Cardiac wall 					& LV septum 		& $3.1 \pm 1.6$  & $3.8 \pm 1.2$ 	& $<0.001$\\ 
								& RV free wall margin & $3.2 \pm 1.4$  & $3.8 \pm 1.1$ 	& $<0.001$\\ 
Cardiac cavity 					& LV trabeculation  & $3.4 \pm 1.5$  & $4.3 \pm 1.1$ 	& $<0.001$\\ 
								& Left arterial appendage 				& $3.0 \pm 1.4$  & $3.6 \pm 0.9$ 	& $<0.001$\\ 
Aorta 							& Aortic root 		& $4.3 \pm 1.2$  & $5.0 \pm 0.5$ 	& $<0.001$\\ 
Valve 							& Aortic valve 		& $2.8 \pm 1.4$  & $3.4 \pm 0.9$ 	& $<0.001$\\ 
								& Mitral valve 		& $2.8 \pm 1.4$  & $3.4 \pm 1.0$ 	& $<0.001$\\  \hline
\end{tabular}}
\end{table*}

\begin{table*}[t!]
\caption{Comparison of standard deviation and signal-to-noise ratio between original CT and denoising CT images measured from selected structures (LV, left ventricle; pRCA, proximal right coronary artery;)}
\label{table4}
\centering
\resizebox{0.6\paperwidth}{!}{
\begin{tabular}{lllllll}
\hline
 		& \multicolumn{2}{l}{Image noise} & \multirow{2}{*}{P-value} & \multicolumn{2}{l}{\multirow{2}{*}{SNR}} & \multirow{2}{*}{P-value} \\
 		& \multicolumn{2}{l}{(standard deviation)} & & \multicolumn{2}{l}{} &   \\ \hline
 				& Original image  & Denoising 		& 		& Original image & Denoising 		& \\ \hline
Ascending aorta & $48.0 \pm 26.8$ & $39.0 \pm 17.0$ & 0.003 & $11.1 \pm 6.5$ & $12.3 \pm 6.2$ 	& 0.001 \\
LV cavity 		& $69.4 \pm 37.9$ & $48.5 \pm 19.0$ & $<0.001$ & $8.9 \pm 12.7$ & $9.0 \pm 3.5$ & 0.96 \\
LV septum 		& $63.2 \pm 40.8$ & $40.9 \pm 14.0$ & $<0.001$ & $2.9 \pm 1.7$ & $3.4 \pm 1.3$ 	& 0.015 \\
pRCA 			& $70.5 \pm 50.7$ & $62.1 \pm 48.5$ & 0.036 & $6.4 \pm 3.9$ & $7.6 \pm 4.4$ 	& 0.034 \\ \hline
\end{tabular}}
\end{table*}

\section{Results}
\label{sec:result}

\subsection{Qualitative evaluation}

To verify the performance of the proposed method, we tested 50 cases from the dataset of mitral valve prolapse patients which were not used in the training session.
Also, we tested 50 cases from the dataset coronary artery disease patients which were not used to training the network.
The results are described in Fig. \ref{fig:result_1} and \ref{fig:result_2}, respectively.
Each row indicates  the different patient case, and the restoration results  from the first column are shown in the second column.
The input low-dose CT images are from phase 1 and the target routine-dose images are from phase 8.
Due to the cardiac motion during CT scanning,  the shape of the heart and image intensity from  the contrast  agent are different at the two phases.
Distinct differences are indicated by the yellow arrows in the images.
The denoised results showed that  the proposed method is good at reducing the noise in the input CT images while the texture information and edges are still intact.
The difference images showed that the proposed method did not change the detailed information and only removes noise from the input CT images.
The proposed method is robust to the type of heart disease as confirmed in another disease cases in  Fig. \ref{fig:result_2}.
Results showed that the network does not create any artificial features  that  can disturb the diagnosis while maintaining the crucial information.

We also observed that  the proposed method is automatically adapted  to the noise levels of the input CT images.
Specifically, there are some data which have similar noise level between phase 1 and phase 8 as shown in Fig. \ref{fig:result_identity}.
If the input CT images have a noise level similar to the CT target images, we have found that the proposed generator $G_{AB}$ does not show any noticeable change, as shown in 
 in Fig. \ref{fig:result_identity}.
These results confirms the proposed generator $G_{AB}$ acts as the identity for the images in the target domain, as shown in \eqref{eq:idour}. 

\begin{figure*}[t!]
\centering
\includegraphics[width=16cm]{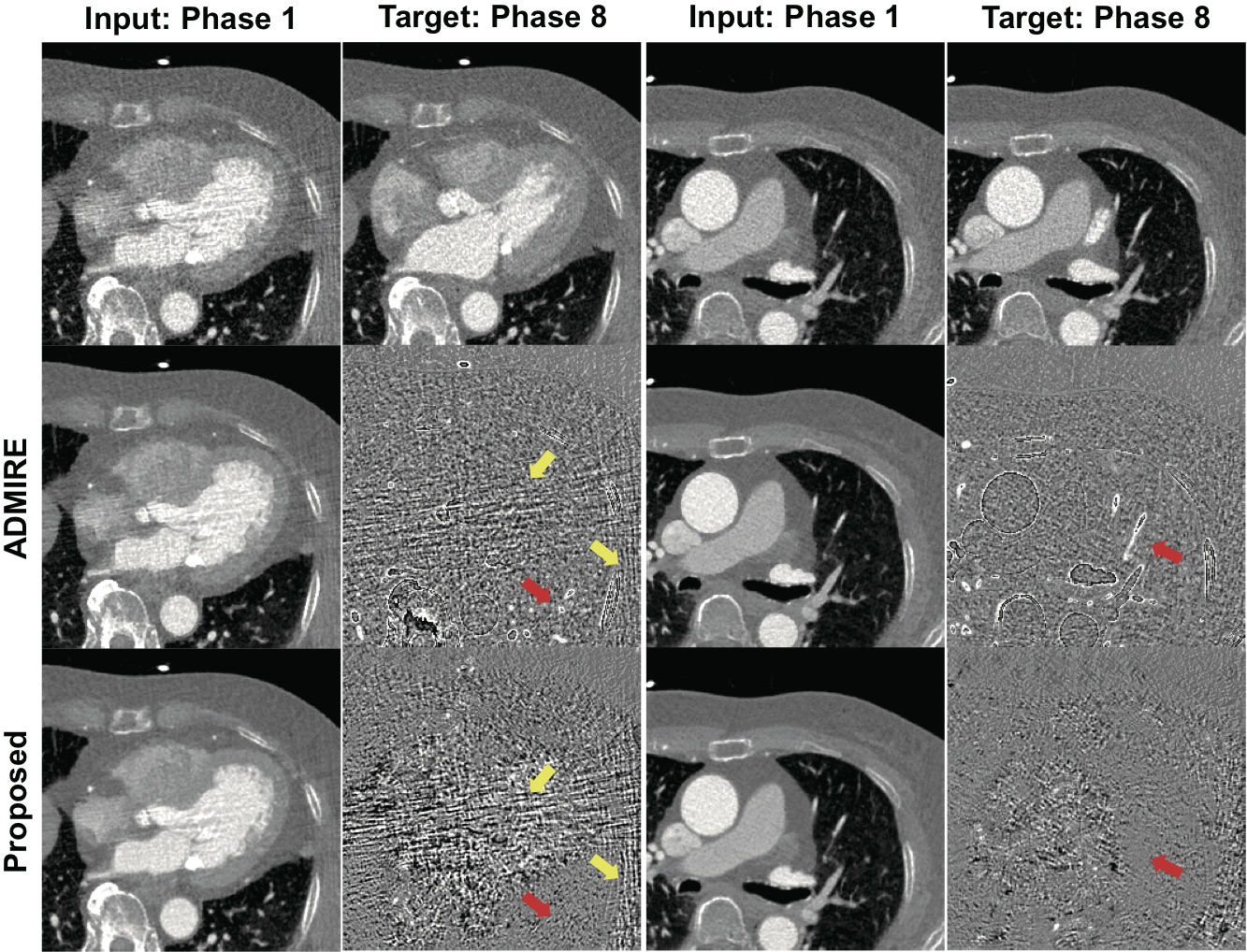}
\caption{Restoration results from the ADMIRE algorithm and the proposed method.
Intensity range of the CT image is (-800, 800)[HU] and the difference image between the input and result is (-100, 100)[HU].
Yellow arrows indicate the streaking noise and red arrows indicate the details in the lung.}
\label{fig:result_admire}
\end{figure*}

To compared the performance with the state-of-the-art model based iterative method (MBIR), 
we compared our algorithm with the Siemens  ADMIRE (Advanced Modeled Iterative Reconstruction)  algorithm \cite{solomon2015diagnostic}.
ADMIRE is the latest MBIR method from Siemens, which has been improved from SAFIRE (Sinogram Affirmed. Iterative Reconstruction) algorithm.
ADMIRE incorporates statistical modeling,
both in the raw projection data
and in the image domains, such that a
different statistical weighting is applied
according to the quality of the projection  \cite{solomon2015diagnostic},
so ADMIRE is only available for latest scanner (Siemens Flash system). Thus,
we cannot provide ADMIRE images for all patients in our retrospective studies, 
so we obtained multiphase CTA images  from a new patient case.

As shown in Fig. \ref{fig:result_admire},
both ADMIRE  and the proposed method successfully  reduced noise in low-dose CT images. However,
the difference images between input and results  showed that, in the case of ADMIRE,  the
edge information was somewhat lost and over-smoothing was observed  in the lung region,  indicated by red arrows. 
On the other hand,  no structural loss was observed in the proposed method.
Moreover,  in the left two columns of Fig. \ref{fig:result_admire}, 
we can clearly see the remaining streaking artifacts in the ADMIRE images, while no such artifacts are observed in the proposed method.
A similar, consistent improvement by the proposed method was observed in all volume slices.

\begin{figure*}[t!]
\centerline{
\includegraphics[width=17cm]{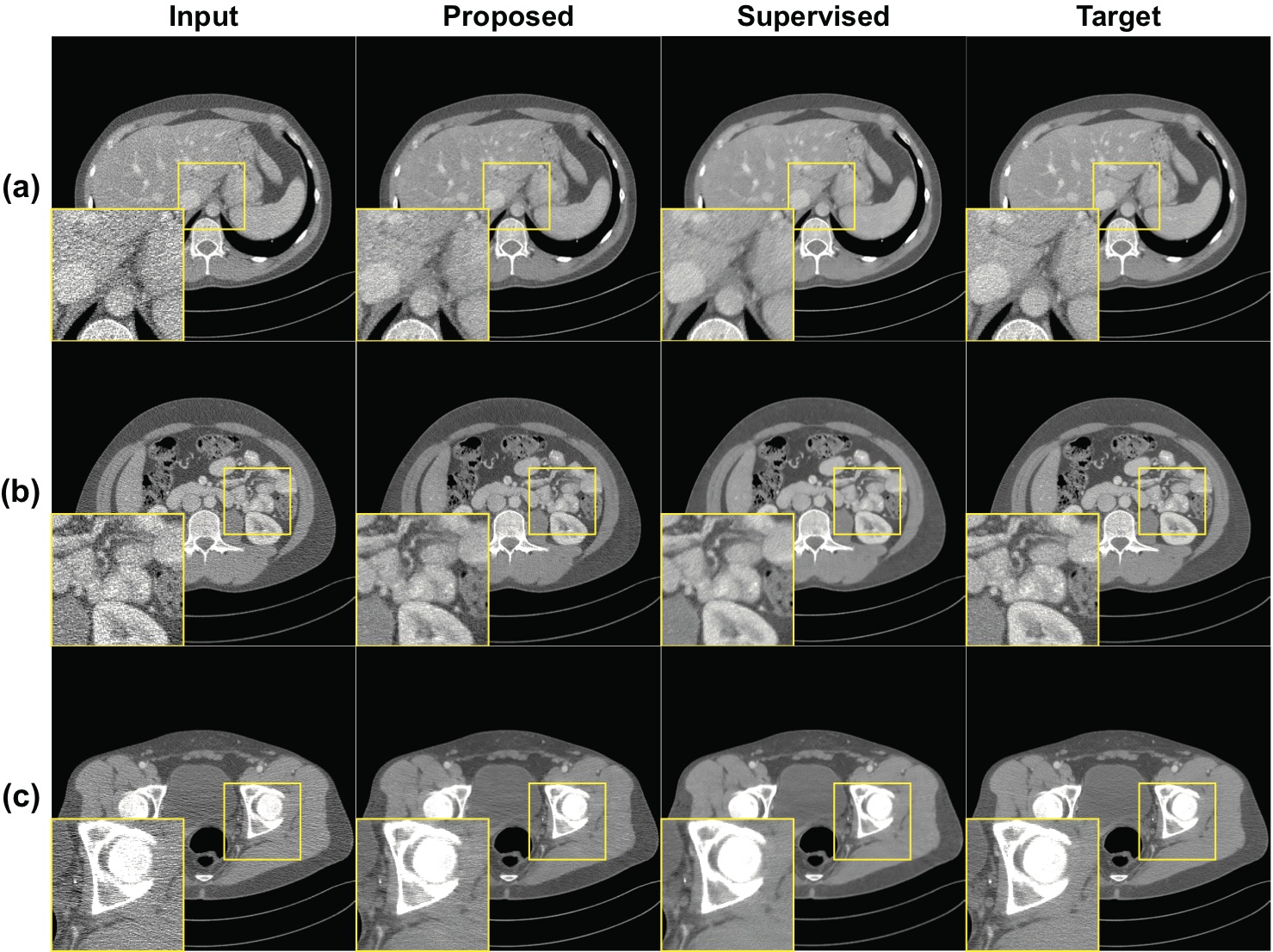}}
\caption{Restoration results from the AAPM challenge dataset using the proposed method and the supervised learning method\cite{kang2017deep}.
Images of (a) the liver, (b) various organs includes the intestine and kidney, etc, (c) the bones.
Intensity range of the CT image is (-300, 300)[HU].}
\label{fig:result_aapm}
\end{figure*}

\subsection{Visual grading score and SNR analysis results}

All visual scores are significantly higher in denoising CT, representing that the image quality of denoising CT is better $(P < 0.001)$ (Table \ref{table3}). 
Quantitatively, image noise was decreased, and SNR was significantly increased on denoising CT $(P <0.05)$ (Table \ref{table4}, Fig. \ref{fig:result_noise_SNR}), 
except no statistically significant SNR changes detected in left ventricular cavity where contrast enhanced blood pool measured by the largest region of interest $(P=0.96)$.

\subsection{Application to AAPM Data Set}

We have performed additional experiments with AAPM  low-dose CT grand challenge dataset
 which consists of abdominal CT images from ten patients.
We used the 8 patient data for training and validation, and the remaining 2 patient data for the test.
In contrast to the existing supervised learning approaches for low dose CT denosing\cite{kang2017deep},
here, the training was conducted in an unsupervised manner using the proposed network, with the input and target images randomly selected from the entire data set.
Fig. \ref{fig:result_aapm}(b) showed that the proposed unsupervised learning method provided even better images than the supervised learning,
while there are some remaining artifacts in Fig. \ref{fig:result_aapm}(a)(c). 
In general, the  denoising results by the proposed approach has the competitive denoising performance compared to the supervised learning approach\cite{kang2017deep}.

\begin{figure*}[t!]
\centering
\includegraphics[width=16cm]{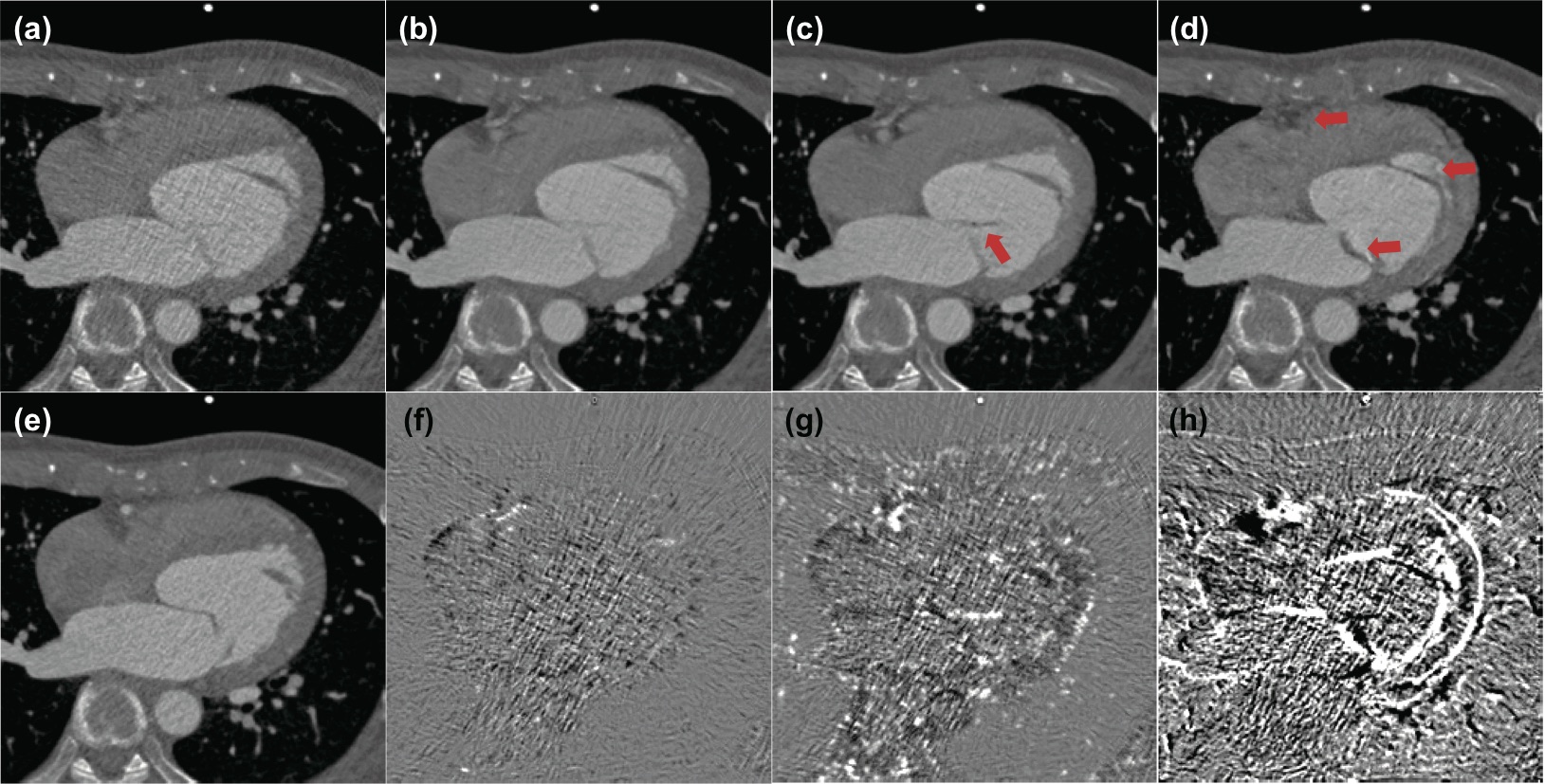}
\caption{(a) Input CT image, (b) proposed method, (c) proposed method without identity loss, (d) with only GAN loss,
(e) target CT image, (f-h) difference images between input image and result images (b-d), respectively.
Intensity range of the CT image is (-820, 1430)[HU] and the difference image between the input and result is (-200, 200)[HU].
Red arrow indicates the artificial features that were not present in the input image.}
\label{fig:result_ablation1}
\end{figure*}

\begin{figure*}[t!]
\centering
\includegraphics[width=16cm]{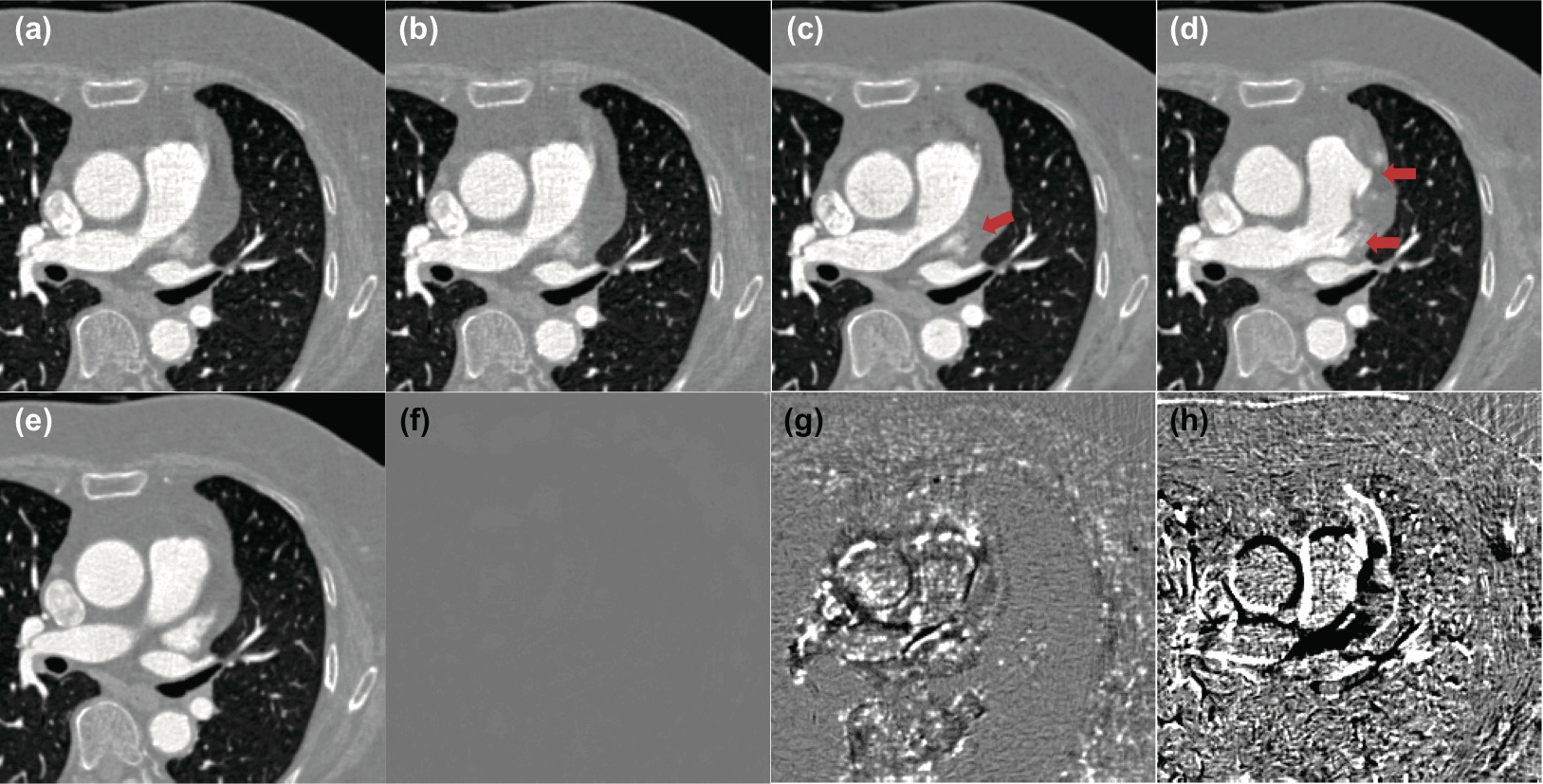}
\caption{(a) Input CT image, (b) proposed method, (c) proposed method without identity loss, (d) with only GAN loss,
(e) target CT image, (f-h) difference images between input image and result images (b-d), respectively.
Intensity range of the CT image is (-924, 576)[HU] and the difference image between the input and result is (-100, 100)[HU].
Red arrow indicates the artificial features that were not present in the input image.}
\label{fig:result_ablation2}
\end{figure*}

\subsection{Ablation study}

\vspace*{-0.2cm}
To analyze the roles of each building block  in the proposed network architecture,
we performed  ablation studies by excluding the identity loss and/or cyclic loss and using the same training procedures.
The results with respect to two different  noise levels are illustrated in Fig. \ref{fig:result_ablation1} and Fig. \ref{fig:result_ablation2}, respectively.
Recall that the input low-dose CT image in the sub-figure (a) and target routine-dose image in the sub-figure (e) have different shape of heart due to the cardiac motion.
The results of the proposed method are illustrated  in the second column, the results of the excluding the identity loss are in the third column, and the results of the excluding the identity loss and cyclic loss are illustrated in the fourth column.
We illustrate reconstruction images as well as the difference images between the input and the reconstruction results.
We also indicate the artificial features that were not present in the input images by red arrows.

All the reconstruction result images in Fig. \ref{fig:result_ablation1} show that the noise level is reduced and  the edge information is well maintained.
In contrast to  the proposed method that does not generate any artificial features, 
the other methods generated some structures which are not present in the input images.
The result of the excluding the identity loss (third column) are better than
the network trained only with GAN loss without including cycle consistency and identity loss (fourth column), 
but both methods deformed the shape of the heart and removed some structures. 
Similar observations can be found in Fig. \ref{fig:result_ablation2} where input CT image has a similar noise level with target CT image.
While the proposed method does not change the original image, the other methods
deformed the shape and created the features that were not present in the input image.
Considering that artificial features can confuse radiologists in diagnosing the patient's disease,  the result confirmed the critical importance of cyclic loss and the identity loss as proposed by our algorithm.

\section{Discussion}

\begin{figure*}[t!]
\centering
\includegraphics[width=8cm]{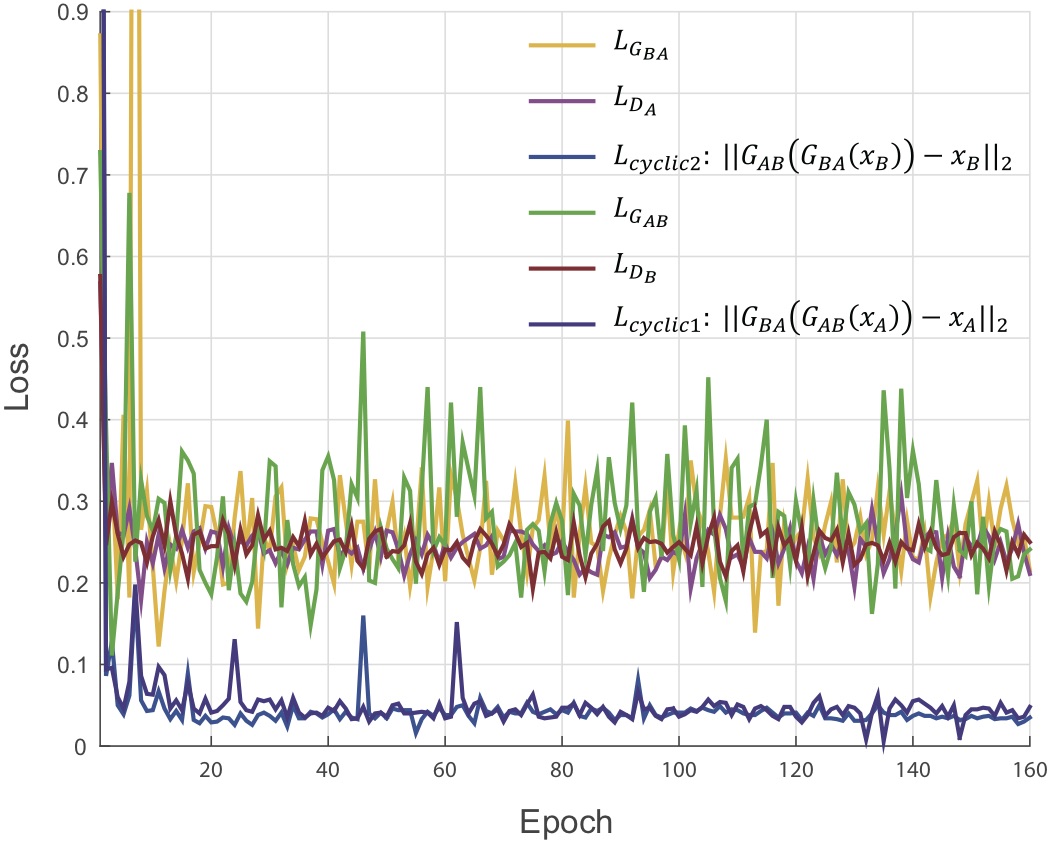}
\caption{Convergence plots according to the epochs during the training process.}
\label{fig:result_convergence}
\end{figure*}

Unsupervised learning with GAN has become popular in computer vision literatuires, which has demonstrated impressive performance for various tasks,
but the classical GAN \cite{goodfellow2014gan} using the sigmoid cross entropy loss function  is often unstable during training.
To address this, we used  LSGAN \cite{mao2017least} and the cycle-loss\cite{zhu2017cyclegan}.
Convergence plots in Fig. \ref{fig:result_convergence} shows that the proposed networks converged stably.
Here, $\Lc_{G_{AB}}$ and  $\Lc_{D_B}$ denoted the loss of generator Eq. \ref{eq:loss_generator} and the loss of discriminator Eq. \ref{eq:loss_discriminator}.
If network reaches the optimal equilibrium state, $\Lc_{G_{AB}}$ and $\Lc_{D_B}$ should be reached at $0.25$, which was also  shown in Fig. \ref{fig:result_convergence}.
The cyclic loss also decreased steadily  during training process and converged. This confirms that the network training was well done.

Another critical issue with GAN is the problem of mode collapse.
The GAN mode collapse occurs when the generator network generates limited outputs or even the same output, regardless of the input.
Unlike the classic GAN, which generates samples from random noise inputs, our network creates samples from the noisy input that are closely related.
In addition, the presence of an inverse path reduces the likelihood of mode collapse, and the identity loss prevents the creation of artificial features. Thanks to the synergistic combination of these components of network architectures,  the likelihood of  mode collapse was significantly reduced, and we have not observed any case where the generated outputs from distinct inputs are the same.

However, there are some limitations of the present studies.
The current method mainly focused on multiphase CTA, and the performance of the proposed method is confirmed in this specific application.
Also, our training, validation, and test data are generated using the same reconstruction kernel (B26f: cardiac filter). 
Thus, it is not clear whether our approach can be generalized to different kernels, organs, etc.
Even though we provided preliminary results using the AAPM data set, more extensive study is required to validate the generalizability of the proposed method.
These issues are very important for clinical uses, which need to be investigated in separate works.

We agree that once  a well-trained network from supervised learning tasks is available, one can use low-dose acquisition for all cardiac phases. However, extensive clinical evaluation is required to have such drastic protocol changes, which is unlikely to happen in the near future. On the other hand, the proposed approach still uses the current acquisition protocols, but provide enhanced images as additional information for radiologists, which can be easily accepted in the current clinical setting.
Moreover,  in contrast to supervised learning approaches for low-dose CT, the  unsupervised learning approaches, such as the proposed one, do not require vendor-supported simulated low-dose data or additional matched full/low-dose acquisition. Therefore, we believe that the potential for the proposed method in terms of science and product development could be significant.

\section{Conclusion}

In this paper, we proposed a cycle consistent adversarial denoising network for multiphase coronary CT angiography.
Unlike the existing supervised deep learning approaches for low-dose CT, our network does not require exactly matched low- and routine- dose images. 
Instead, our network was designed to learn the image distributions from the high-dose cardiac phases.
Furthermore, in contrast to the other state-of-the-art deep neural networks with GAN loss that are prone to generate artificial features, 
our network was designed to prevent from generating artificial features that are not present in the input image by exploiting the cyclic consistency and identity loss.
Experimental results confirmed that the proposed method is good at reducing the noise in the input low-dose CT images while maintaining the texture and edge information.
Moreover, when the routine dose images were used as input, the proposed network did not change the images, confirming that the algorithm correctly learn the noise.
Radiological evaluation using visual grading analysis scores also confirmed that
the proposed denoising method significantly increases the diagnostic quality of the images.
Considering the effectiveness and practicability of the proposed method,
our method can be widely applied for other CT  acquisition protocols  with dynamic tube current modulation.

\section*{Acknowledgement}

The authors would like to thanks Dr. Cynthia MaCollough, the Mayo Clinic, the American Association of Physicists in Medicine (AAPM), and grant EB01705 and EB01785 from the National Institute of Biomedical Imaging and Bioengineering for providing the Low-Dose CT Grand Challenge data set.
This work is supported by Industrial Strategic technology development program (10072064, Development of Novel Artificial Intelligence Technologies To Assist Imaging Diagnosis of Pulmonary, Hepatic, and Cardiac Disease and Their Integration into Commercial Clinical PACS Platforms) funded by the Ministry of Trade Industry and Energy (MI, Korea).
This work is also supported by the R\&D Convergence Program of NST (National Research Council of Science
\& Technology) of Republic of Korea (Grant CAP-13-3-KERI).


%

\end{document}